\documentclass[letterpaper]{article}
\usepackage{imav}
\usepackage{times}
\usepackage{graphicx}

\usepackage{amsmath}
\usepackage{amssymb}
\usepackage{hyperref}
\usepackage{url}
\usepackage[nolist,nohyperlinks,printonlyused]{acronym}
\usepackage[nointegrals]{wasysym}

\newcommand{\R}{\mathbb{R}}
\newcommand{\degrees}{^\circ}

\title{Flying a Quadrotor with Unknown Actuators and Sensor Configuration}
\author{T.M. Blaha%
\thanks{This work has been submitted to IMAV 2024 for possible publication.
Copyright may be transferred without notice, after which this version may
no longer be accessible.\\
Email address(es): \{t.m.blaha; e.j.j.smeur; b.d.w.remes; c.c.devisser\}@tudelft.nl\\
Data: \href{https://doi.org/10.4121/0530be90-cc6c-4029-9774-670657882906}{10.4121/0530be90-cc6c-4029-9774-670657882906}\\
Code: \href{https://github.com/tudelft/indiflightSupport/tree/iros_imav_2024}{github.com/tudelft/indiflightSupport/tree/iros\_imav\_2024}}, E.J.J. Smeur, B.D.W. Remes and C.C. de Visser \\ Delft University of Technology, Kluyverweg 1, 2629 HS Delft, The Netherlands}

\begin{document}

\begin{acronym}
\acro{ML}{Machine Learning}
\acro{INDI}{Incremental Nonlinear Dynamic Inversion}
\acro{NDI}{Nonlinear Dynamic Inversion}
\acro{UAV}{Unmanned Air Vehicle}
\acro{RLS}{Recursive Least Squares}
\acro{LS}{Least Squares}
\acro{LMS}{Least Mean Squares}
\acro{ESC}{Electronic Speed Control}
\acro{RMS}{Root-Mean-Square}
\acro{EKF}{Extended Kalman Filter}
\acro{IMU}{Inertial Measurement Unit}
\acro{MRAC}{Model-Reference Adaptive Control}
\acro{MEMS}{Micro Electro-Mechanical System}
\acro{CoG}{Center of Gravity}
\acro{UKF}{Unscented Kalman Filter}
\acro{GNSS}{Global Navigation}
\end{acronym}

\maketitle
\thispagestyle{empty} 

\begin{abstract}
Though control algorithms for multirotor \ac{UAV} are well understood, the configuration, parameter estimation, and tuning of flight control algorithms takes quite some time and resources.
In previous work, we have shown that it is possible to identify the control effectiveness and motor dynamics of a multirotor fast enough for it to recover to a stable hover after being thrown 4 meters in the air.
In this paper, we extend this to include estimation of the position of the \ac{IMU} relative to the \ac{CoG}, estimation of the \ac{IMU} rotation, the thrust direction of all motors and the optimal combined thrust direction.
In order to guarantee a correct \ac{IMU} position estimation, two prior throw-and-catches of the vehicle with spin around different axes are required. For these throws, a height as low as 1 meter is sufficient.
Quadrotor flight experimentation confirms the efficacy of the approach, and a simulation shows its applicability to fully-actuated crafts with multiple possible hover orientations.
\end{abstract}

\acresetall

\section{Introduction} \label{section:introduction}

Drones are used for an ever increasing number of applications, with variations in the number and placement of propellers, the craft's inertia and sensor location and orientation.
Careful controller design is needed to ensure new platforms are stable and demonstrate good flight performance.
However, the process of developing, testing, and fine-tuning flight control algorithms for new drones is both time-consuming and expensive, which poses a significant barrier to the advancement of new drone applications.

In recent work \cite{blaha_control_2024}, we have shown that it is possible to learn all required control parameters in the span of a single throw, such that a multirotor learns to fly without prior knowledge of its motor dynamics time constant, actuator position and effectiveness.
In that paper, two important assumptions were made.
First it was assumed that the motors were providing upward thrust in a body frame, simplifying the position control to that of a standard multirotor.
The motors may not always be mounted in this configuration, or in the case of fully or overactuated multirotors, independent thrust can be generated in any direction, allowing multiple hover attitudes \cite{rajappa_modeling_2015}.
The second assumption was that the \ac{IMU} orientation and position offset relative to the \ac{CoG} were known.
This assumption was needed, because of the high angular rates the vehicle obtains during throwing and excitation, which necessitate compensation of the acceleration measured by the \ac{IMU} to transform it to the \ac{CoG}.

\begin{figure}[t]
    \centering
    \includegraphics[width=0.8\columnwidth]{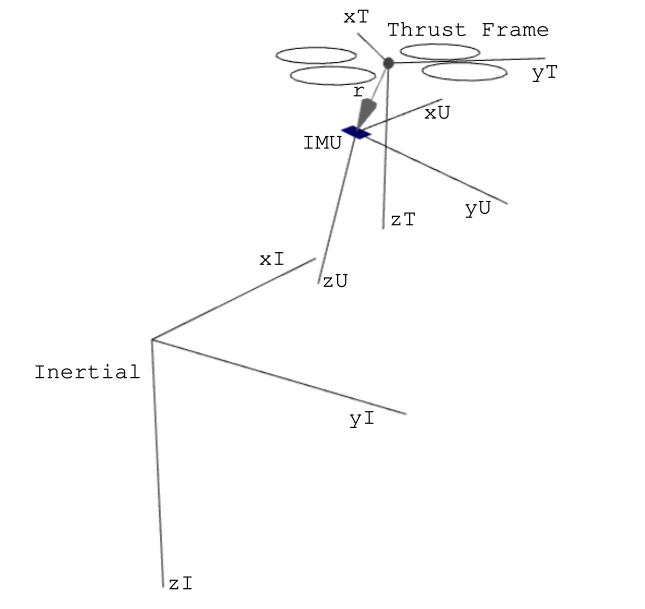}
    \caption{Coordinate Systems: Identifying the transformation between thrust frame $T$ and IMU frame $U$ from less than $1$ second of flight data is the main contribution of this work.}
    \label{fig:frames}
\end{figure}

Online estimation of the position of the \ac{CoG} relative to the \ac{IMU} was addressed in \cite{fresk_generalized_2017}, but this was done relying on actuator inputs with a predetermined effectiveness. Svacha et al. \cite{svacha_imu-based_2020} implemented a 32 state \ac{UKF} estimating all mass properties, including \ac{IMU} offset. However, this was done with offline data, with known mass and motor locations. Lukacs and Lantos \cite{lukacs_nonlinear_2015} provide an algorithm for determining the position of the \ac{CoG} together with a force model for a sailplane, but practical results are not provided.
\cite{burri_maximum_2016} estimated the moments of inertia, \ac{CoG} position, and aerodynamic parameters of a quadrotor model using a Minimum Likelihood Estimator, but this can only be run offline.
The typical approach in the literature seems to be to estimate the \ac{CoG} offset at the same time as the control effectiveness, which either requires some assumptions on the geometry of the drone, or extends the estimation time significantly.



In this paper, we extend the work of \cite{blaha_control_2024} to unknown \ac{IMU} orientation and position, and unknown thrust directions of the actuators. However, the basic assumption that the vehicle has to be capable of static hover is maintained.
The only user input to the system is two throw-and-catches of about 1 meter height with sufficient spin and one throw-and-fly of around 3.5 meter height.
The applicability of the algorithm is shown through real life experiments with a quadrotor.




\section{Frames of Reference} \label{section:frames}

We distinguish 3 coordinate frames, illustrated in Figure \ref{fig:frames}, and relate them using unit quaternions.
The frame of the \ac{IMU} with respect the inertial frame is described by $q^U_I$.
The "thrust frame" $T$ is defined with its origin in the center of gravity of the vehicle, and the $z^T$-axis pointing opposite a chosen thrust axis.
This axis is obvious for quadrotors, but e.g. for fully actuated hexarotors that can hover in many different orientations\cite{rajappa_modeling_2015}, the choice may not be obvious, so that we define $-z^T$ as the axis of \emph{most efficient} thrust generation capacity while simultaneously not causing any body-moments.
The choice of $x^T$ is commonly taken as the front of the vehicle, but in our study this choice is arbitrary.

The thrust frame relates to the IMU frame with $q^T_U$, and to the inertial frame with the intrinsic rotation sequence $q^T_I = q^U_I q^T_U$.

A problem with the arbitrary definition of the $x^T$ and $y^T$ axes, is that the yaw angle needs to be known in order to control the position and velocity of a multirotor.
We see two solutions for this problem.

The first is an \ac{EKF} that estimates the heading by fusing \ac{GNSS} measurements and \ac{IMU} acceleration measurements projected onto the inertial xy-plane.
The second is to include a magnetometer, of which the axes are aligned with the \ac{IMU}.


In this study we only consider \ac{IMU} rotations around an axis in the thrust-xy plane.
This guarantees that a desired inertial accelerations can be translated to the correct attitude changes for closed loop control.

\section{Methodology}

This section gives a brief overview of the state estimation, controller structure and system identification onboard of the UAV. Then our method to determine the IMU location and the thrust frame is explained with focus on real-time implementation.

\subsection{Controller Summary and Effectiveness identification}

The same controller and estimator structure of \cite{blaha_control_2024} is also used in this work. A short summary:

A complementary attitude filter is used to estimate $q^U_I$. The yaw angle computed by an external motion capture system is fused into this estimate.

An \ac{INDI} controller is used to give motor commands to achieve a desired $q^T_I$, which in turn is computed by a \ac{NDI} position controller.

The identification of the effectiveness matrix and motor model for the INDI controller results from a \ac{RLS} implementation based on the measured angular rate derivatives and specific forces in response to a sequential open-loop excitation of each motor.


\subsection{IMU location estimation}\label{sec:IMUestimation}

The inertial acceleration $a_r$ of a point $r$ expressed in the coordinates of a local, rotating frame $T$ is given by:
\begin{equation}\label{eq:accel}
    a_r = a_T + \ddot r + 2\Omega_T \times \dot r + \dot\Omega_T \times r + \Omega_T \times (\Omega_T \times r)\ ,
\end{equation}

where $a_T$ is the inertial acceleration of the frame, $r,\ \dot r,\ \ddot r$ is the location, velocity and acceleration of $r$ relative to frame $T$, and $\Omega_T,\ \dot\Omega_T$ the rotation rate and acceleration of frame $T$. All quantities are expressed in the coordinates of frame $T$.

A rigidly mounted \ac{IMU} is now placed at $r$, and the origin of the rotating frame is chosen at the vehicle's center of gravity: $\dot r = \ddot r = 0$.
Since accelerometers do not measure gravity (but rather all other specific forces), its reading for a free-tumbling rigid body on which no external forces act will follow this reduced equation with $a_T = 0$:

\begin{align}\label{eq:accelSimplified}
    a_r &= \dot\Omega_T \times r + \Omega_T \times (\Omega_T \times r)\\
    \underbrace{a_r}_{y_i} &= \underbrace{ \begin{pmatrix}
        -\Omega_{y}^{2} - \Omega_{z}^{2} & \Omega_{x} \Omega_{y} - \dot{\Omega}_{z} & \Omega_{x} \Omega_{z} + \dot{\Omega}_{y} \\
        \Omega_{x} \Omega_{y} + \dot{\Omega}_{z} & -\Omega_{x}^{2} - \Omega_{z}^{2} & \Omega_{y} \Omega_{z} - \dot{\Omega}_{x} \\
        \Omega_{x} \Omega_{z} - \dot{\Omega}_{y} & \Omega_{y} \Omega_{z} + \dot{\Omega}_{x} & -\Omega_{x}^{2} - \Omega_{y}^{2}
    \end{pmatrix} }_{X_i}
    \underbrace{
    \begin{pmatrix}
        r_x \\ r_y \\ r_z    
    \end{pmatrix}
    }_{r}.
\end{align}

This equation is linear in the parameters $r$, and so if $\Omega,\ \dot\Omega,\ a_r$ are measured during such a tree-tumble, the observations $X_i$ and $y_i$ can be aggregated in matrix $X$ and vector $y$. \ac{LS} can then be used to obtain an estimate of $r$.

The conditions of a tumbling rigid body with minimal external forces (e.g. from aerodynamic drag) can be achieved by throwing it in the air with sufficient angular velocity.
One single throw of the vehicle would be ideal, however it is generally not enough to reliably fit $r$: if the vehicle is rotating around a single axis (without loss of generality, assume the axis $(0,\ 0,\ 1)$), then the resulting equation is 

\begin{align}
    a_r =
    \begin{pmatrix}
        \Omega_z^2 & -\dot\Omega_z & + \dot\Omega_y \\
        \dot\Omega_z & -\Omega_z^2 & - \dot\Omega_x \\
        -\dot\Omega_y & \dot\Omega_x & 0
    \end{pmatrix}
    \begin{pmatrix}
        r_x \\ r_y \\ r_z    
    \end{pmatrix}\ , 
\end{align}

and so if $(0,\ 0,\ 1)$ happens to be close to a principle inertial axis, then $\dot\Omega$ is expected to be small and $r_z$ cannot be estimated.
It is later experimentally illustrated that two throws about significantly different axes provide enough excitation.

\subsection{Thrust Direction Estimation} \label{section:thrustDirection}

Instead of estimating the \ac{IMU}'s orientation with respect to some other body-fixed frame, we estimate the direction in which force can be generated with the motors, without producing body-moments.
For conventional quadrotors there is only one such direction (coinciding with the axes of each motor), but many more directions are possible for configurations with more actuators and tilted rotors. If we knew a suitable such direction (e.g. the most power-efficient one), we can use this to construct a reference frame in which a conventional multirotor controller can operate.

Recall that we assume the multirotor configuration to be able to achieve steady-state hover. We now approximate the specific forces and torques acting on the vehicle as linear combinations of the motor input $u\in \R^m$, where $m$ is the number of motors on the vehicle:

\begin{equation}\label{eq:steadystate}
    \begin{pmatrix}f\\\tau\end{pmatrix} = \begin{pmatrix}G_{1,f}\\G_{1,\tau}\end{pmatrix}u\ .
\end{equation}

Here, the $f\in \R^3$ and $\tau \in \R^3$ are the specific forces and torques, respectively, and $G_{1,f} \in \R^{3\times m}$ and $G_{1,\tau} \in \R^{3\times m}$ are the corresponding steady-state effectiveness matrices.
In \cite{blaha_control_2024}, these matrices are identified locally using the increments of $u$ and $(f,\ \tau)^T$, but here we assume that they hold in the global sense of Equations \ref{eq:steadystate}.
Note that Equation \ref{eq:steadystate} is only valid in steady-state when $\dot u = 0$.

The most efficient distribution of control effort in hover can be defined as the one that minimizes some square norm of $u$, while ensuring (1) there are no torques acting on the vehicle ($\tau=0$), (2) the gravity is canceled by the forces ($||f||=g$), and (3) the bounds on $u$ are respected.
Introducing the diagonal actuator weighing matrix $W$, these statements are summarized in the non-convex quadratically constrained quadratic problem

\begin{align} \label{eq:QCQP}
\begin{split}
    \min_u\quad u^T W u \\
    s.t.\quad u^T (G_{1,f}^T G_{1,f}) u &= g^2 \\
    \quad G_{1,\tau}u &= 0\\
    \underline{u} \leq u &\leq \overline{u}\ .
\end{split}
\end{align}

If a solution $u=u^*$ exists, then the vehicle is indeed capable of static hover and the most efficient direction to generate thrust without causing rotations is $d = G_{1,f}u^*$.

General quadratic programming solvers readily solve this problem, however, their implementation in embedded hardware is difficult, and computation time is slow and convergence time may not be guaranteed.
If $G_1$ changes, or at least its estimate, then Problem \ref{eq:QCQP} would have to be solved again and again, which quickly becomes infeasible.

\paragraph{Real-Time Implementation}

We now introduce a heuristic and a simplification that allows development of an efficient algorithm to solve Problem \ref{eq:QCQP} in sufficient cases.

Firstly, assume that the bounds $\underline{u} \leq u \leq \overline{u}$ are always fulfilled if the first two constraints are met, and can thus be dropped.
This may cause the found solution to be infeasible if some motors are at their limit or not used at all at true solution $u^*$. The rationale for this simplification is that it is unlikely that a vehicle capable of static hover requires motors at maximum or minimum thrust even in its most efficient hovering attitude.

Secondly, assuming identical motors, impose $W=I$. This will cause sub-optimal hover distributions when the motors do not have the same power consumption for a given input $u$.

The algorithm follows these steps
\begin{enumerate}
    \item Introduce new variables $\eta\in\R^{m-3}$ by calculating an orthogonal basis $N_\tau$ for the $(m-3)$-dimensional nullspace of $G_{1,\tau}$ (full-rank is guaranteed by the assumption that the vehicle can achieve static hover).
    This can be efficiently achieved by a column-pivoting complete QR decomposition of $G_{1,\tau}^T$, e.g. with the LAPACK routine \textsc{SGEQP3}\footnote{\href{https://netlib.org/lapack/explore-html/d0/dea/group__geqp3_gab537b8a58f9a711c429f72ff9483d668.html}{netlib.org/lapack/explore-html/d0/dea/group\_\_geqp3}}.
    
    \item Now, any choice of $u = N_\tau \eta$ satisfies the constraint $G_{1,\tau}u = 0$, and what remains is 
    \begin{align}\label{eq:reduced}
    \begin{split}
        \min_\eta\quad \eta^T N_\tau^T W N_\tau \eta \\
        s.t.\quad \eta^T (N_\tau^T G_{1,f}^T G_{1,f}N_\tau) \eta &= g^2\ .
    \end{split}
    \end{align}
    Defining $H\triangleq N_\tau^T W N_\tau$, and $A \triangleq N_\tau^T G_{1,f}^T G_{1,f}N_\tau$, and introducing the Lagrange multiplier $\lambda$, we can write the KKT system of Equation \ref{eq:reduced} as 
    \begin{align}\label{eq:KKT}
    \begin{split}
        2H\eta + 2\lambda\cdot A\eta &= 0\\
        \eta^TA\eta &= g^2
    \end{split}
    \end{align}
    
    \item It is now clear why we imposed $W=I$: $N_\tau$ is an orthogonal basis for the nullspace, due to our choice of the QR factorization. Consequently, $H=I$ as well, and the first line of Equation \ref{eq:KKT} is an eigen-problem $Av_i = \tilde{\lambda}_i v_i$, where we explicitly, but without loss of generality assume $||v_i|| = 1$. 
    Now we know that $\eta^* = k v_i$ with one of the eigenvectors of $A$ and some scaling factor $k$, which can be computed by substituting this relation into the second line of \ref{eq:KKT}:
    \begin{align}
        k^2 v_i^T \underbrace{A v_i}_{\tilde{\lambda}_i v_i} = g^2  \implies  k = \frac{g}{\sqrt{\tilde\lambda_i v_i^T v_i}} = \frac{g}{\sqrt{\tilde\lambda_i}} \ .
    \end{align}

    Consequently, $u^* = \pm N_\tau \eta^* = \pm N_\tau k v_i = \pm N_\tau g / \sqrt{\tilde\lambda_i} v_i$, where the sign must be chosen so that the components of $u^*$ are within $\underline{u} \leq u \leq \overline{u}$.
    Following our heuristic, if this is impossible, then the vehicle is very likely not capable of static hover in the first place. This relation also shows that we require not just any, but the largest eigen-pair (or one of the largest eigen-pairs, if there are multiple equal ones) of $A$, as that results in the lowest magnitude of $u^*$. 

    \item Since $A\succ0$, we can compute the largest eigen-pair of $A$ using the power iteration $v_{j+1} = \frac{Av_j}{||Av_j||}$. In practice an iteration limit should be imposed, and since the parameters of the problem are not expected to change rapidly, subsequent invocations of step 4. will continue to improve the guess. 

    \item Finally, the axis of most efficient thrust generation is then $d = G_{1,f} u^*$, where $u^* = \pm N_\tau g / \sqrt{\tilde\lambda} v_i$, with $\pm$ and $v_i$ as determined above. Then, $q^T_U$ can finally be assembled as the shortest rotation that maps $d$ to $(0,\ 0,\ -g)$.
\end{enumerate}

Most of the computation time of the algorithm above is expected in the computation of the QR-decomposition which increases linearly with the amount of motors $m^2$. The subsequent matrix calculations increase with $(m-3)^2$.

\subsection{Experiment Design -- Quadrotor}

\begin{figure}
    \centering
    \includegraphics[width=0.75\linewidth]{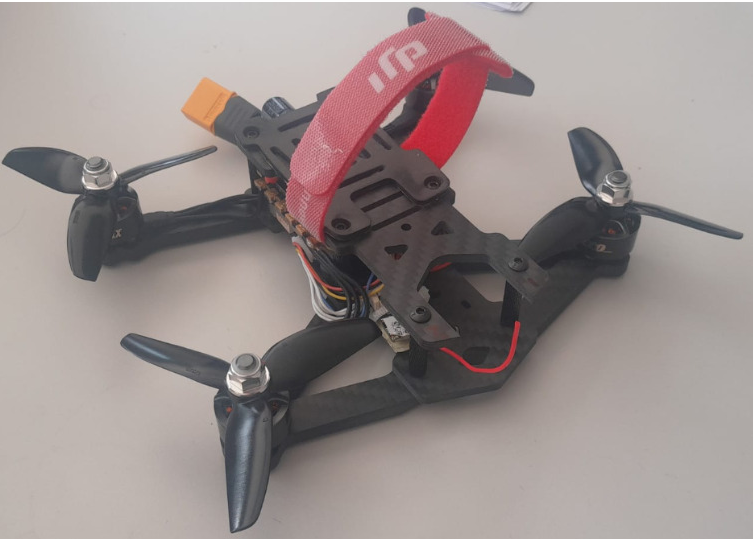}
    \caption{Compact $3$-inch quadrotor used for the flight experiments.}
    \label{fig:cinerat}
\end{figure}

The estimation and control algorithm is verified experimentally, in the same fashion as done in \cite{blaha_control_2024}, by implementing the computations in the firmware INDIflight\footnote{\href{https://github.com/tudelft/indiflight}{github.com/tudelft/indiflight}} onboard of an STM32H743 flight controller (480 MHz clockspeed), and using gyroscope and accelerometer data from a TDK InvenSense ICM-42688-P IMU. The vehicle is pictured in Figure \ref{fig:cinerat}.
The update rate of the attitude control and actuator effectiveness estimation is $2kHz$. Even though it is not necessary to run the thrust frame estimation at the same rate, this is still performed for ease of comparison of the runtimes later.
For safety reasons, the quadrotor is launched and given an initial rotation of $\Omega_0 = (400,\ 400,\ 100)^T$ deg/s to simulate a human throw. The invariance of the estimations to this initial condition has been shown in \cite{blaha_control_2024}.

In order to test different \ac{IMU} orientations without making changes to the hardware, the \ac{IMU} measurements are rotated by the inverse of $q^T_U = 0.837 - 0.491i + 0.242j$, which corresponds to Euler angles of  $-15^\circ$ Yaw, $24^\circ$ Pitch, $-64^\circ$. As explained in the limitation of Section \ref{section:frames}, this rotation results in correct yaw alignment when rotated back using the algorithm of Section \ref{section:thrustDirection}.

Though it is entirely possible to do the estimation of the IMU location $r$ onboard, for the demonstration of this work it has been performed offline with logged sensor measurements, and was then manually programmed into the memory of the flight controller. Onboard estimation of the parameter would merely require detection of the 2 required throws and casting the \acf{LS} estimator as a \acf{RLS} estimator to eliminate the memory requirements of the offline \ac{LS}.

\begin{figure}[t]
    \centering
    \includegraphics[width=0.8\columnwidth]{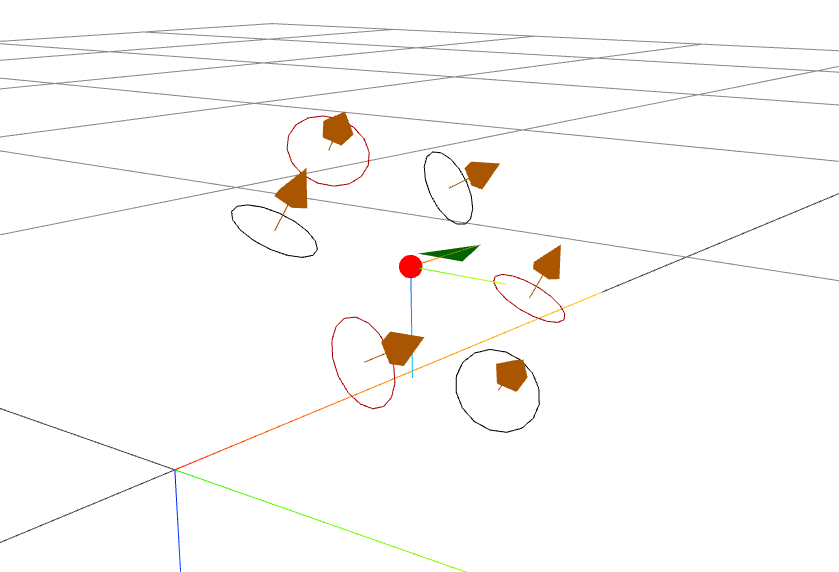}
    \caption{Fully-actuated hexarotor that can hover in multiple orientations.}
    \label{fig:hexarotor}
\end{figure}

\subsection{Simulation Design -- Fully Actuated Hexarotor}

Steps 3 through 5 of the algorithm in Section \ref{section:thrustDirection} only need to involve the matrix algebra if the rotational nullspace $N_\tau$ spans multiple dimensions. In the case of a quadrotor, dim$\left(N_\tau\right) = 1$ and steps 3-5 are trivial.

The hexarotor (dim$\left(N_\tau\right)=3$) shown in Figure \ref{fig:hexarotor} has an arrangement such that there is one energy-optimal hover orientation, obvious by inspection, but many other suboptimal ones \cite{rajappa_modeling_2015}. To test our algorithm, a simulation model has been created with the added difficulty that the entire actuator constellation is rotated by $45\degrees$ with respect to the IMU (the green arrow in Figure \ref{fig:hexarotor}).
This illustrates again the equivalence of unknown IMU orientation, and unknown actuator locations; our approach does not need an explicitly defined ``body-frame''.

The simulation is run with the same control code as in the flight-experiment, and may be reproduced using Docker\footnote{\href{https://github.com/tudelft/indiflightSupport/tree/iros_imav_2024}{github.com/tudelft/indiflightSupport/tree/iros\_imav\_2024}}.

\section{Results}

All data used in this work is available for reproduction in a dataset\footnote{\href{https://doi.org/10.4121/0530be90-cc6c-4029-9774-670657882906}{10.4121/0530be90-cc6c-4029-9774-670657882906}}.

\begin{figure}[t]
    \centering
    \includegraphics[trim={0 2.5cm 0 0cm}, clip, width=\columnwidth]{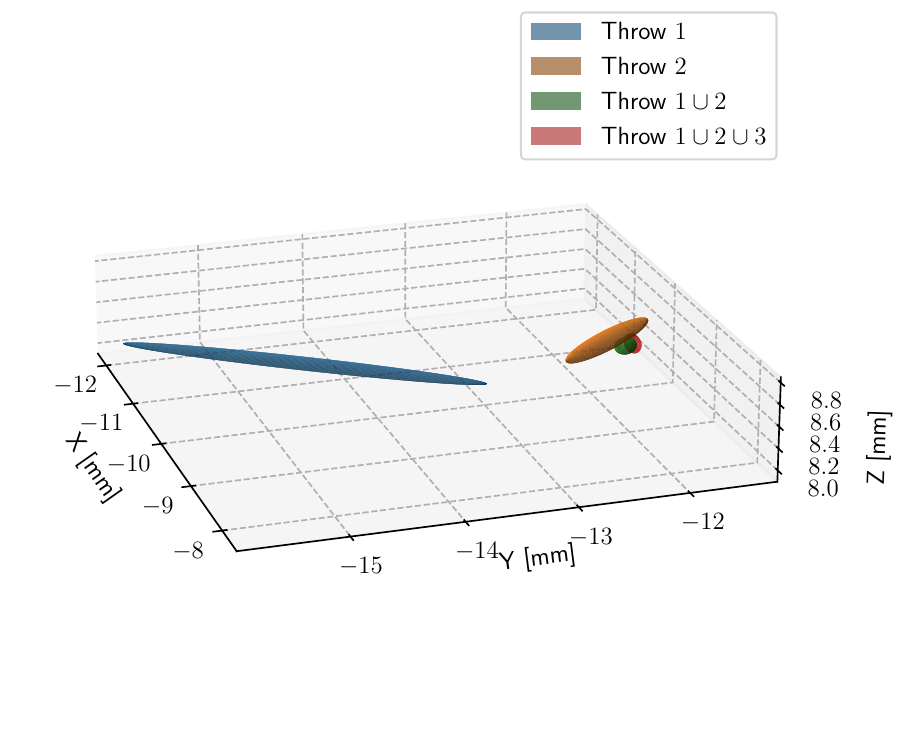}
    \caption{95\% percent confidence ellipsoids of the IMU offset parameter for a sequence of throws, where $1\cup 2$ stands for the union of the data obtained from throws 1 and 2. 
    Least Squares fails for just 1 throw, but succeeds for multiple.}
    \label{figure:imuEllipsoids}
\end{figure}

\subsection{IMU location}

Figure \ref{figure:imuEllipsoids} shows the $95\%$ percentile confidence ellipsoids for the parameters $\hat{r}$ after \ac{LS} fitting, defined by $(r - \hat{r})^T \Sigma^{-1} (r - \hat{r}) < iCDF(\chi_3^2, 0.95)$. Here, $\Sigma \approx \text{SE}^2 \cdot (X^T X)^{-1}$ and the standard error of the residuals $\text{SE}^2 = ||y - X\hat{r}||_2^2 / (N-3)$ \cite{wasserman_lecture_2017}. 

These ellipsoids capture the effects of correlation in the problem data $X$ and the residual error variance in the predictions $y - X\hat{r}$. Systematic errors like those arising from, e.g. inaccurate scaling of the IMU measurements, are not captured.

Using only the observations from a single throw yields uncertain estimates in one direction, as indicated in Section \ref{sec:IMUestimation}.
However, combining two throws with spin around two different axes reduces the confidence bounds to the sub-millimeter range and is not much improved by adding data from a third throw. The remaining uncertainty may be due to sensor bias, aerodynamic drag being a factor ($a^T \neq 0$), and the individual test masses of the 3 \ac{MEMS} accelerometers not being exactly in the same location.



\subsection{Flight Experiment Results -- Quadrotor}


\begin{figure*}[t]
    \centering
    \includegraphics[trim={0mm 0 0mm 0}, clip, width=0.63\linewidth]{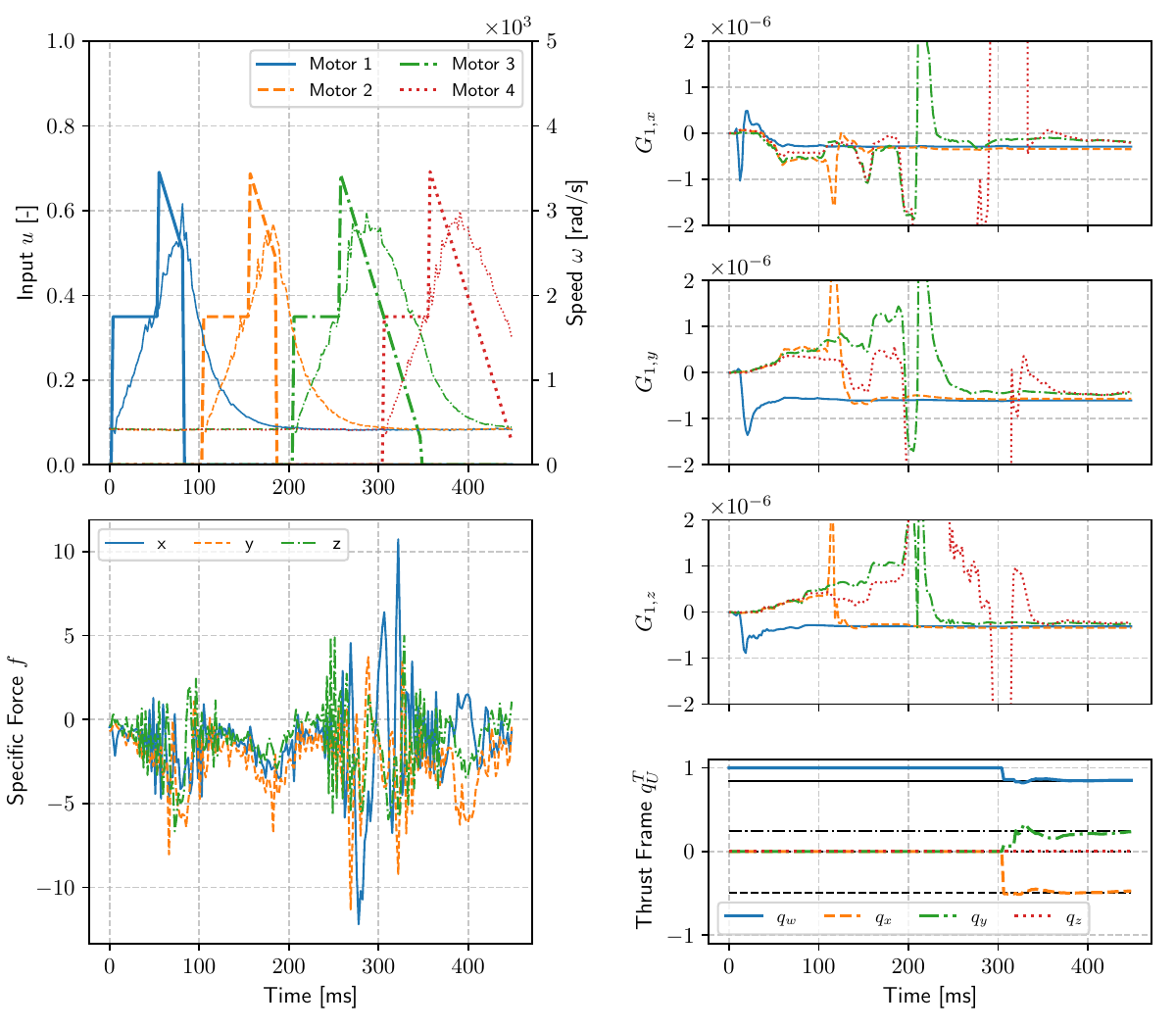}
    \caption{Flight experiment: Inputs $u$, motor response $\omega$, accelerometer measurements $f$ and fitting of the linear part of the Effectiveness Matrix ($G_{1,xyz}$). The last plot shows that the true IMU alignment is approached closely, once the last motor is excited.}
    \label{figure:hoverAttitude}
    \vspace{-5mm}
\end{figure*}
\begin{figure*}[!h]
    \centering
    \includegraphics[trim={0mm 0 0mm 0}, clip, width=0.63\linewidth]{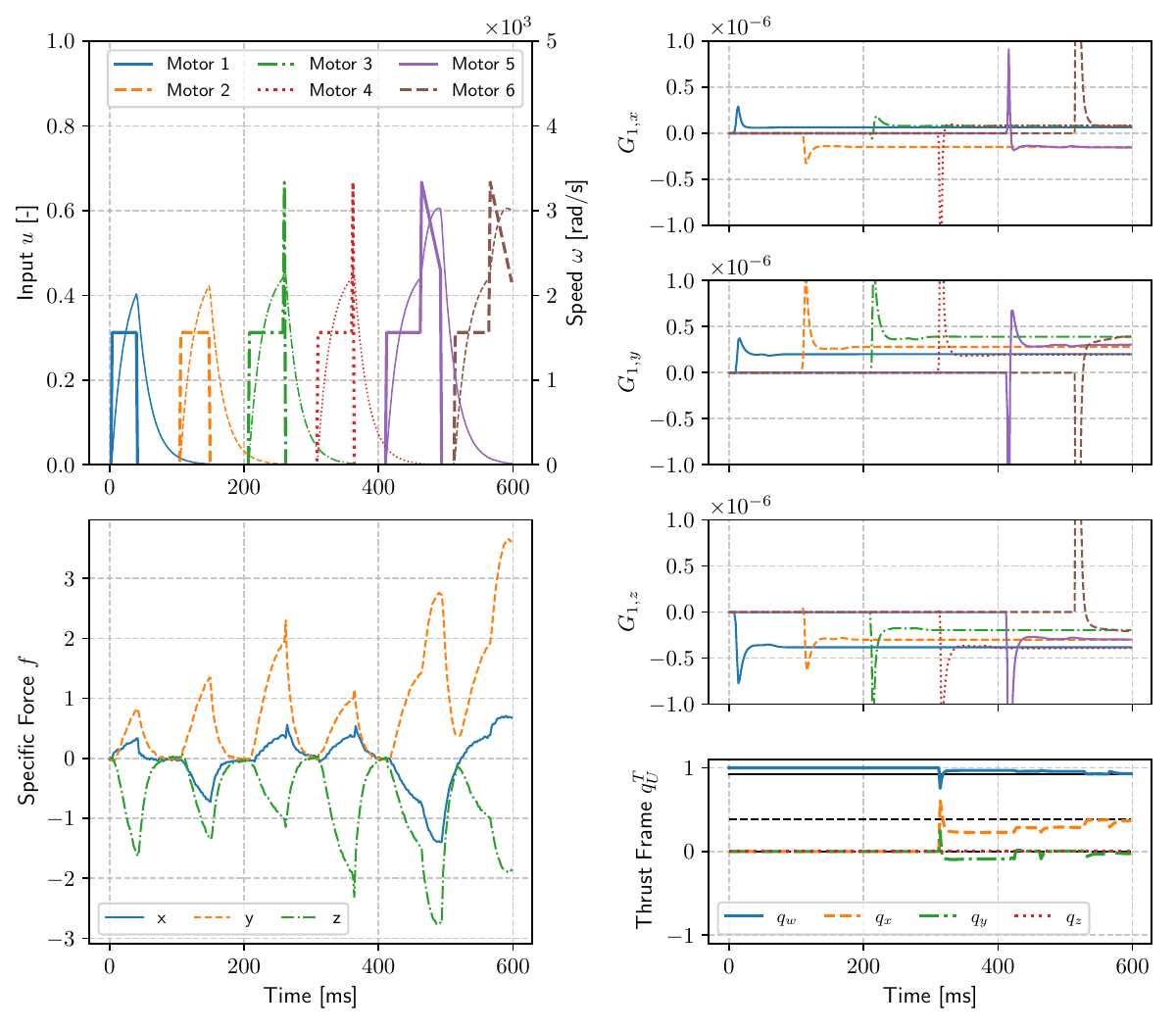}
    \caption{Hexarotor simulation: The parameters of the 6 actuators are identified in the span of 600ms. The specific-force part of $G_1$ (xyz-components shown on the top right for all 6 actuators) show the significant contributions in all IMU axes, not just $z$. From these measurements, our method estimates the optimal thrust frame.}
    \label{figure:hoverAttitudeSim}
    \vspace{-5mm}
\end{figure*}

The recovery of the vehicle and attaining a position setpoint follows a similar trajectory to the results shown in \cite{blaha_control_2024}.
The excitation and system response are shown on the left in Figure \ref{figure:hoverAttitude}. Due to the alignment of the IMU, each motor gives rise to specific forces in all three axes.
The right side of Figure \ref{figure:hoverAttitude} shows the identification of the translational part of the effectiveness matrix $G_{1,f}$, where every vertical dashed line corresponds to the start of the excitation of the next actuator.
Once all actuators have been triggered (roughly $300$ms into the sequence), the computation of the algorithm determining the thrust direction (Section \ref{section:thrustDirection}) runs to completion, as $G_{1,\tau}$ now has a nullspace of dimension greater than $0$.

The estimated and true alignment of the IMU is given in Table \ref{tab:imu_rotation}.
The alignment is sufficiently accurate to not affect manual or automatic position control performance.

\begin{table}[hbt]
    \centering
\begin{tabular}{c||c|c}
     & Ground truth & Estimated \\
     \hline
    Roll & -15$^\circ$ & -14.1$^\circ$\\
    Pitch & 24$^\circ$ & 23.7$^\circ$\\
    Yaw & -64$^\circ$ & -61.0$^\circ$
\end{tabular}
    \caption{Quadrotor Flight Experiment: Estimated rotation of the IMU compared to the ground truth.}
    \label{tab:imu_rotation}
\end{table}


\subsection{Execution Timings}

\begin{figure}[h]
    \centering
    \includegraphics[width=\columnwidth]{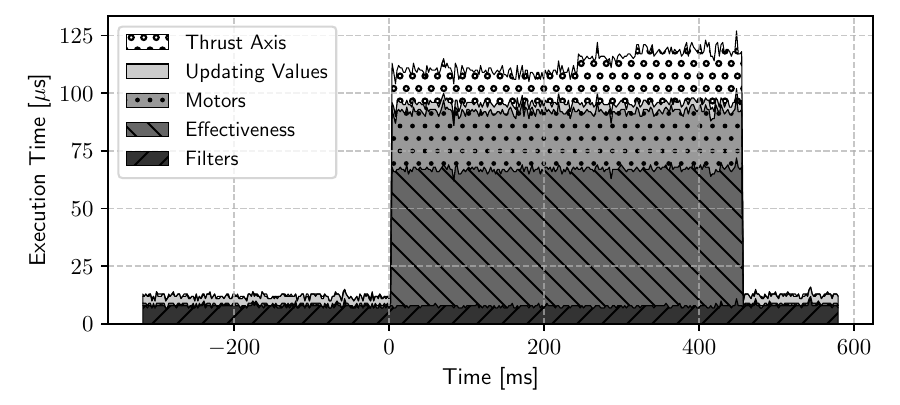}
    \caption{Cumulative execution time on the flight controller show that $2$kHz update rates are feasible for quadrotors.}
    \label{figure:execTimes}
\end{figure}

All execution on the Flight Controller is single-thread, which means that execution time can be measured accurately, with only interrupt handling being a source of error. Figure \ref{figure:execTimes} shows the cumulative execution times of the components of the system identification and thrust axis determination algorithms.
The compilation was done with \texttt{gcc 10.3.1} at optimization level \texttt{-O3}.

The thrust axis estimation adds only a small delay, amounting to an additional $2kHz \cdot 20e-6 s = 4\%$ CPU load, over the other identification components present in \cite{blaha_control_2024}. 

\subsection{Simulation Results -- Hexarotor}

Also in the simulated throw-to-hover, the results show that the optimal thrust frame is estimated correctly at the end of the fitting process.
In the ground truth, this frame of most efficient hover has been rotated in roll by $45\degrees$, and so the attitude quaternion that our algorithm seeks is  $q^T_U = 0.924 + 0.383i$. 
The lower right subplot in Figure \ref{figure:hoverAttitudeSim} shows that this is reached at the time the last actuator has been excited.
In particular, our method found $\hat q^T_U = 0.930 + 0.367i - 0.026j$, which is shown in Euler angles in Table \ref{tab:imu_rotation_sim}.

\begin{table}[hbt]
    \centering
\begin{tabular}{c||c|c}
     & Ground truth & Estimated \\
     \hline
    Roll & 45$^\circ$ & 43.1$^\circ$\\
    Pitch & 0$^\circ$ & -2.74$^\circ$\\
    Yaw & 0$^\circ$ & -1.08$^\circ$
\end{tabular}
    \caption{Hexacopter Simulation: Estimated rotation of the IMU compared to the ground truth.}
    \label{tab:imu_rotation_sim}
\end{table}

\section{Conclusion}

In this study, we successfully demonstrated attitude control of a quadrotor where the motors' location and strength are unknown, as well as the alignment and location of the IMU. The location of the IMU has been estimated from two prior throw-and-catches, after which the quadrotor was able to recover from being launched on a $3.5$m parabolic trajectory.

Recursive Least Squares was used to identify the motors' rotational and linear effectiveness in-flight, which in turn were used to calculate an optimal thrust direction in the frame of the IMU which can be actuated without giving rise to rotations.
The latter is the main contribution of this paper and enables position control if an external source of position and heading (GPS+magnetometer, optical motion capture) is available and the alignment of the IMU $q^T_U$ is restricted to yaw-free rotations. The presented algorithm computes in $20\mu$s on consumer-grade flight control hardware, however computation time will grow with $(m-3)^2$, where $m$ is the number of actuators.

The ability of our method to compute the most efficient hover direction, as opposed to just any, was confirmed in a simulation of a fully actuated hexarotor.

In future work, position control should be enabled even in the cases where the IMU-to-hover frame alignment includes yaw components, either by including a magnetometer sensor aligned with the IMU, or by calculating heading from comparing the vehicles GPS-track with the acceleration measured onboard.

Furthermore, the dependence on prior throws may be eliminated by eliminating the need to know the IMU location before identification of the motors' effectiveness.
This could be achieved by providing additional excitation in the nullspace of $G_{1,\tau}$ once rotations have been arrested.
In this case, the accelerometer will measure the true specific forces, as every point on the vehicle experiences the same acceleration.

\section*{Acknowledgements}
The authors would like to thank Elijah Hao Wei Ang for his insights.



\appendix
\newcommand{\appsection}[1]{\let\oldthesection\thesection
  \renewcommand{\thesection}{Appendix \oldthesection:}
  \section{#1}\let\thesection\oldthesection}

\end{document}